\documentclass[11pt]{article}

\usepackage[final]{acl}

\usepackage{times}
\usepackage{latexsym}
\usepackage{amssymb}
\usepackage{amsmath}
\usepackage{tcolorbox} 
\usepackage{tabularx}  
\usepackage[ruled,vlined,linesnumbered]{algorithm2e}
\DontPrintSemicolon
\usepackage{booktabs}
\usepackage{multirow}
\usepackage{makecell}
\usepackage{caption}
\usepackage{float}
\usepackage{geometry}
\usepackage[T1]{fontenc}

\usepackage[utf8]{inputenc}

\usepackage{microtype}

\usepackage{inconsolata}

\usepackage{graphicx}

\title{E2E-GMNER: End-to-End Generative Grounded Multimodal Named Entity Recognition}

\author{
  \textbf{Meng Zhang\textsuperscript{1}},
  \textbf{Jinzhong Ning\textsuperscript{1}\thanks{Corresponding authors.}},
  \textbf{Xiaolong Wu\textsuperscript{1}},
  \textbf{Hongfei Lin\textsuperscript{2}},
  \textbf{Yijia Zhang\textsuperscript{1}\footnotemark[1]} \\
  \textsuperscript{1}Dalian Maritime University \\
  \textsuperscript{2}Dalian University of Technology
}

\begin{document}
\maketitle
\begin{abstract}
Grounded Multimodal Named Entity Recognition (GMNER) aims to jointly identify named entity mentions in text, predict their semantic types, and ground each entity to a corresponding visual region in an associated image. Existing approaches predominantly adopt pipeline-based architectures that decouple textual entity recognition and visual grounding, leading to error accumulation and suboptimal joint optimization. In this paper, we propose \textbf{E2E-GMNER}, a fully end-to-end generative framework that unifies entity recognition, semantic typing, visual grounding, and implicit knowledge reasoning within a single multimodal large language model. We formulate GMNER as an instruction-tuned conditional generation task and incorporate chain-of-thought reasoning to enable the model to adaptively determine when visual evidence or background knowledge is informative, reducing reliance on noisy cues. To further address the instability of generative bounding box prediction, we introduce Gaussian Risk-Aware Box Perturbation (GRBP), which replaces hard box supervision with probabilistically perturbed soft targets to improve robustness against annotation noise and discretization errors. Extensive experiments on the Twitter-GMNER and Twitter-FMNERG benchmarks demonstrate that E2E-GMNER achieves highly competitive performance compared with state of the art methods, validating the effectiveness of unified end-to-end optimization and noise-aware grounding supervision. Code is available at: \url{https://github.com/Finch-coder/E2E-GMNER}
\end{abstract}

\section{Introduction}
Grounded Multimodal Named Entity Recognition (GMNER) is a vision–language task that jointly identifies named entity mentions in text, predicts their semantic types, and grounds each entity to its corresponding visual region in the associated image. By providing explicit entity-level alignments between text and visual evidence, GMNER enables structured multimodal understanding that supports downstream applications such as multimodal knowledge graph construction \cite{mmkg} and visually grounded question answering \cite{VQA}.

\begin{figure}[t]
  \centering
  \includegraphics[width=\linewidth]{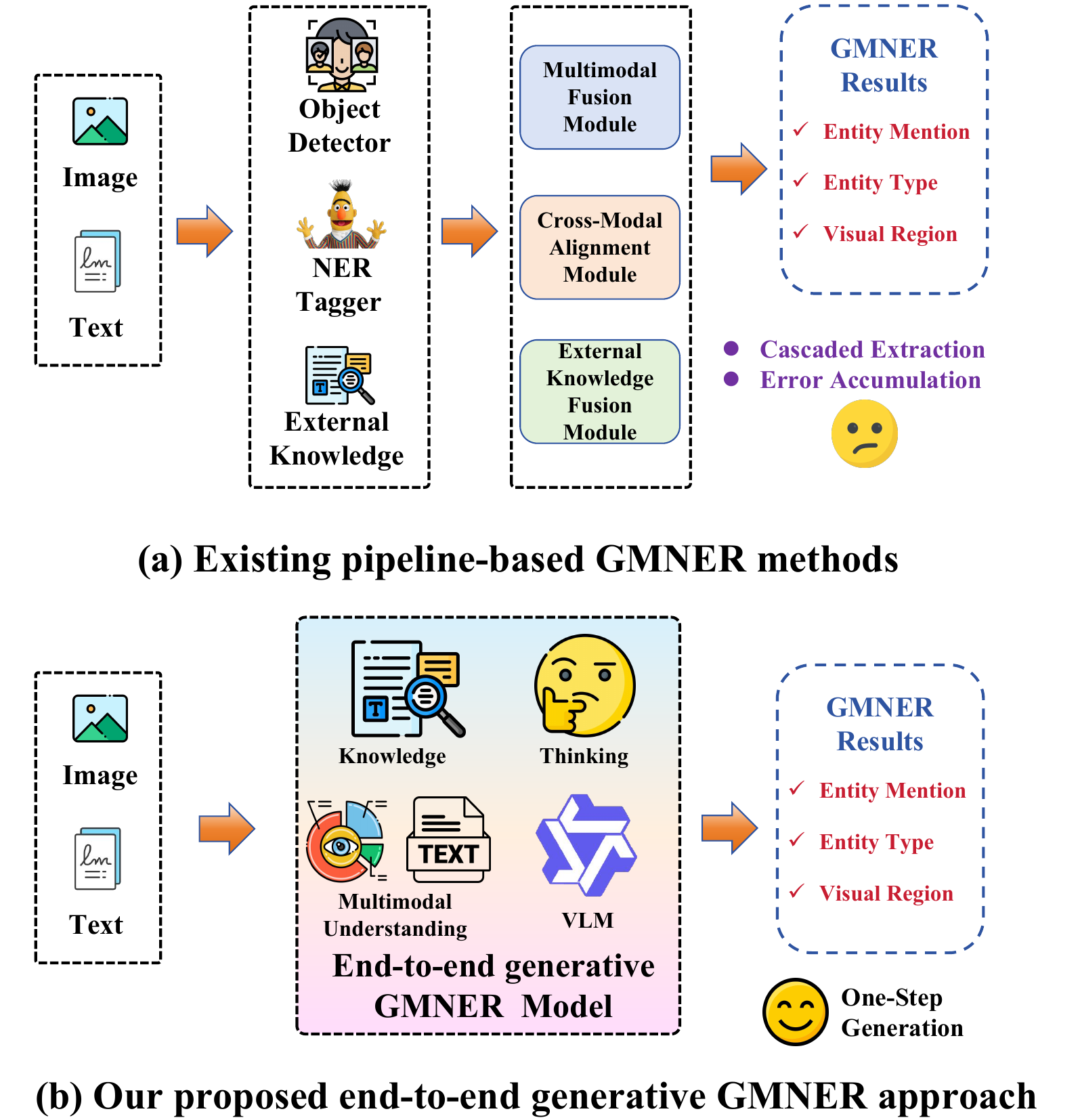}
  \caption{Comparison between existing pipeline-based GMNER methods and our end-to-end generative GMNER approach.}
  \label{figure1}
\end{figure}

Most existing GMNER approaches adopt a pipeline-based architecture to perform cascaded extraction of named entities and their corresponding visual targets. Specifically, they typically rely on a separate BERT-based sequence labeling NER model \cite{makar,RiVEG} to identify entity mentions in text, or employ external object detectors \cite{GMNER,FGMNER,unco} to extract region proposals from images, followed by a fusion module that aggregates textual and visual representations to produce final GMNER predictions. In addition, some methods incorporate external knowledge sources, such as knowledge bases \cite{scanner,makar} or knowledge generated by prompting large language models \cite{MVP,GEM}, to assist joint text–image understanding.

Despite recent advances, GMNER poses several fundamental challenges.
\textbf{Issue1: }Most methods adopt a pipeline architecture that decouples textual entity recognition and visual grounding, relying on separate modules such as standalone NER taggers or external object detectors, whose predictions are cascaded to produce the final GMNER outputs. This design leads to error accumulation across stages and prevents joint optimization.
\textbf{Issue2: }Although most approaches attempt to resolve text–vision ambiguity through implicit cross-modal alignment or by incorporating external knowledge cues, they often lack an explicit mechanism to determine when visual evidence or external knowledge is truly informative. As a result, disambiguation can be suboptimal when visual cues are noisy, irrelevant, or misleading.
\textbf{Issue3: }Under emerging generative vision–language grounding paradigms \cite{qwen2.5vl,kosmos}, directly predicting bounding boxes as sequences of discrete coordinate tokens introduces additional training challenges. Supervision based on a single hard target sequence makes learning sensitive to annotation noise and discretization errors, thereby reducing robustness and optimization stability.

To address the above limitations, we propose \textbf{E2E-GMNER}, a novel end-to-end framework for GMNER that enables unified optimization of entity recognition, visual grounding, semantic alignment, and background knowledge reasoning within a single generative vision-language model, as illustrated in \textbf{Figure~}\ref{figure1}.
For \textbf{Issue 1}, our framework unifies entity recognition and visual grounding within a single generative formulation, enabling the model to jointly predict entity mentions, their semantic types, and corresponding visual regions. By eliminating intermediate pipeline components, our approach mitigates error accumulation inherent in cascaded GMNER systems and allows joint optimization of entity recognition and grounding.
For \textbf{Issue 2}, we adopt Chain-of-Thought (CoT) instruction tuning to adapt the planning, multimodal semantic reasoning, and knowledge utilization capabilities of vision–language models to the GMNER task. This design enables the model to autonomously decide when visual evidence or external knowledge cues are informative, thereby reducing noise introduced by irrelevant visual or knowledge signals while avoiding explicit reliance on external knowledge sources.
For \textbf{Issue 3}, we introduce Gaussian Risk-Aware Box Perturbation (GRBP) during training. Specifically, we probabilistically perturb the center coordinates and scale of ground-truth bounding boxes, replacing a single hard supervision target with Gaussian-based soft supervision, where larger perturbations are assigned lower probabilities. This strategy improves the robustness of generative box prediction by tolerating small geometric deviations in bounding box coordinates, thereby stabilizing training under annotation noise and discretization effects.

Our contributions are summarized as follows:
\begin{itemize}
    \item We propose E2E-GMNER, an end-to-end generative framework that unifies entity recognition and visual grounding within a single formulation, enabling joint optimization and avoiding error accumulation from pipeline-based GMNER methods.
    \item We incorporate Chain-of-Thought instruction tuning to adaptively decide when visual evidence or external knowledge is informative, and introduce Gaussian Risk-Aware Box Perturbation to stabilize generative box prediction under annotation noise and discretization errors.
    \item Extensive experiments on standard GMNER benchmarks show that E2E-GMNER achieves highly competitive results compared with state-of-the-art methods.
\end{itemize}

\section{Related Work}
\textbf{Grounded Multimodal Named Entity Recognition (GMNER)} aims to extract textual entities, their semantic types, and the corresponding visual regions from image--text pairs. Early approaches, such as H-Index \cite{GMNER} and TIGER \cite{FGMNER}, rely on external object detectors to generate candidate regions and then employ generative models to predict entity mentions and align them with the extracted visual proposals. With the advent of multimodal large language models (MLLMs), methods such as RiVEG \cite{RiVEG} and GEM \cite{GEM} integrate MLLMs to leverage their strong multimodal semantic understanding and knowledge capabilities for grounding entities to visual regions. Recent works including SCANNER \cite{scanner} and UnCo \cite{unco} further improve generalization by incorporating external knowledge, particularly for handling unseen entities, while MQSPN \cite{MQSPN2025AAAI} adopts a set prediction paradigm to alleviate exposure bias in generative GMNER settings. In addition, MAKAR \cite{makar} employs an MLLM-based multi-agent system to resolve semantic ambiguity and enhance alignment between textual entities and visual regions.

It is worth noting that nearly all existing GMNER methods follow a pipeline-based architecture, which decouples textual entity recognition and visual grounding into separate modules such as standalone NER taggers or external object detectors, with their predictions cascaded to produce final GMNER outputs. This design inevitably leads to error accumulation and hinders joint optimization across tasks. \textbf{To the best of our knowledge, our proposed approach is the first fully end-to-end GMNER framework, enabling unified optimization of entity recognition, visual grounding, semantic alignment, and background knowledge reasoning within a single generative model.}

\section{Method}

\begin{figure*}[t]
  \centering
  \includegraphics[width=0.85\linewidth]{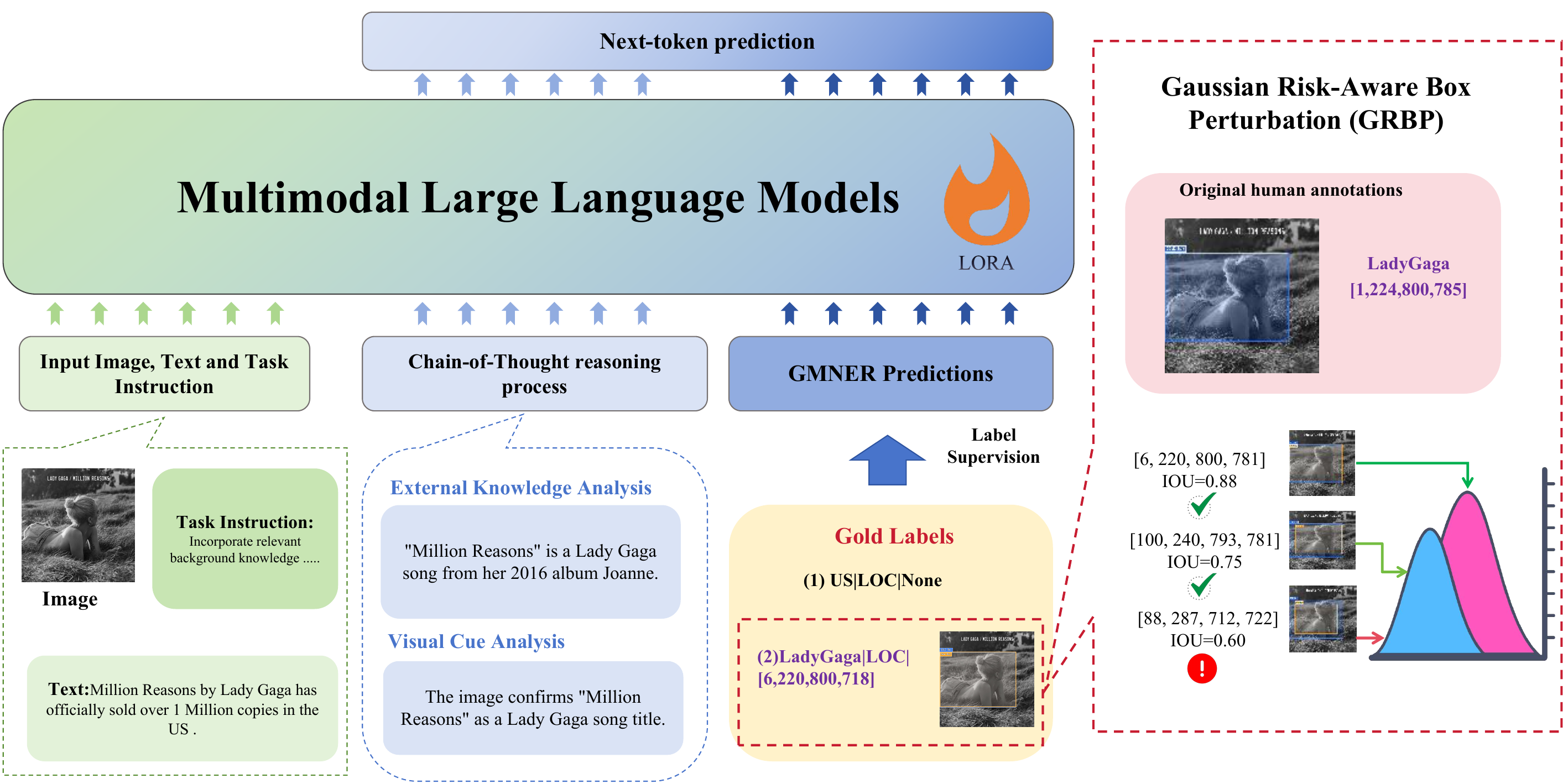}
    \caption{Overview of \textbf{E2E-GMNER}. Given an image, text, and task instruction, a LoRA-adapted multimodal large language model performs chain-of-thought multimodal reasoning (visual cue analysis and background knowledge analysis) and generates GMNER outputs---entity mentions, semantic types, and grounded bounding boxes---via next-token prediction. During training, gold entity-type and box annotations provide supervision, and we further introduce \textbf{Gaussian Risk-Aware Box Perturbation (GRBP)} to replace hard box targets with Gaussian-weighted soft supervision by probabilistically perturbing the ground-truth box (larger perturbations receive lower probability), improving robustness to annotation noise and discretization effects.}
  \label{figure2}
\end{figure*}

\subsection{Task Formulation}

Grounded Multimodal Named Entity Recognition (GMNER) aims to jointly identify named entity mentions in text, predict their semantic types, and ground each entity to a corresponding visual region in the associated image. Formally, given an image--text pair $(I, T)$, where $I$ denotes the image and $T$ denotes the textual sequence, the goal of GMNER is to produce a set of structured entity records:
\begin{equation}
\mathcal{Y} = \{ (e_i, c_i, b_i) \}_{i=1}^{N}
\end{equation}
where $e_i$ is span of the $i$-th entity mentioned in $T$, $c_i$ is its semantic type, and $b_i = (x_i^{1}, y_i^{1}, x_i^{2}, y_i^{2})$ denotes the bounding box grounding the entity in the image $I$, specified by the top-left and bottom-right coordinates. The output set $\mathcal{Y}$ is unordered, and the number of entities $N$ varies across instances. 

\subsection{End-to-End Generative Framework for GMNER}
We propose an end-to-end generative framework for Grounded Multimodal Named Entity Recognition (GMNER) based on a multimodal large language model (MLLM) \cite{qwen2.5vl}. Instead of decomposing the task into separate stages such as textual NER and Object Detection, our framework formulates GMNER as a structured generation problem conditioned jointly on the image and text. The MLLM encodes multimodal inputs into a shared representation and directly generates complete entity records in a single pass, enabling joint optimization of entity recognition, semantic typing, and visual grounding while avoiding error propagation in pipeline-based methods. An overview of the framework is illustrated in \textbf{Figure~}\ref{figure2}.

\subsubsection{Instruction-Tuned Generative Formulation with Chain-of-Thought}

We formulate Grounded Multimodal Named Entity Recognition (GMNER) as an instruction-tuned conditional generation task based on a multimodal large language model (MLLM). Given an image–text pair $(I, T)$, the model input is constructed by prepending a task-specific instruction to the multimodal input:
\begin{equation}
\text{Input} = [\text{Instruction}; (I, T)]
\end{equation}

The MLLM generates outputs autoregressively in a single sequence composed of two parts: a reasoning sequence $R$ and a sequence of structured entity records corresponding to the entity set $\mathcal{Y} = {(e_i, c_i, b_i)}_{i=1}^{N}$. Each entity record is serialized using the following output schema:
\begin{equation}
e_i \mid c_i \mid [x_i^{1},, y_i^{1},, x_i^{2},, y_i^{2}]
\label{eq:output_schema}
\end{equation}
where $e_i$ denotes the entity span, $c_i$ its semantic type, and $[x_i^{1}, y_i^{1}, x_i^{2}, y_i^{2}]$ the grounding bounding box. All entity records are concatenated sequentially to form the final structured prediction.

To enable the model to adaptively determine whether visual evidence or implicit knowledge is informative for accurate grounding, we incorporate chain-of-thought (CoT) reasoning into the generation process. This design allows the model to avoid indiscriminate reliance on potentially noisy visual cues, while facilitating the joint optimization of entity recognition, visual grounding, and implicit knowledge reasoning within a unified generative framework.

Formally, the generation process is expressed as:
\begin{equation}
[R; \mathcal{Y}] = f_{\theta}([\text{Instruction}; (I, T)])
\end{equation}
where $f_{\theta}$ denotes the MLLM. 

During training, reasoning sequences $R$ are provided as supervision signals generated by a stronger external large language model via API-based inference, enabling effective chain-of-thought instruction tuning. At inference time, no external models are used, and the MLLM autonomously generates both reasoning and structured predictions in an end-to-end manner.

\begin{algorithm}[t]
\caption{GRBP: IoU-Guarded Gaussian Box Perturbation}
\label{algorithm1}

\KwIn{Ground-truth box \(b=(x_1,y_1,x_2,y_2)\), image size \((W,H)\), jitter \(\beta,\gamma\), IoU threshold \(\tau\), max tries \(T\), min size \(m\), scale bounds \([s_{\min}, s_{\max}]\).}
\KwOut{Perturbed box \(\tilde b\).}

\((c_x,c_y,w,h) \leftarrow \mathrm{ToCenterSize}(b)\)\;

\If{\(w<m\) \textbf{or} \(h<m\)}{
  \Return \(b\)\;
}

\For{\(t \leftarrow 1\) \KwTo \(T\)}{
  \tcp{Center jitter (relative to box size)}
  draw \(\delta_x,\delta_y \sim \mathcal{N}(0,\beta^2)\)\;
  \(c_x' \leftarrow c_x + \delta_x \cdot w\)\;
  \(c_y' \leftarrow c_y + \delta_y \cdot h\)\;

  \tcp{Scale jitter (multiplicative, bounded)}
  draw \(\epsilon_w,\epsilon_h \sim \mathcal{N}(0,\gamma^2)\)\;
  \(a_w \leftarrow \min(s_{\max},\max(s_{\min},1+\epsilon_w))\)\;
  \(a_h \leftarrow \min(s_{\max},\max(s_{\min},1+\epsilon_h))\)\;
  \(w' \leftarrow \max(m, w \cdot a_w)\)\;
  \(h' \leftarrow \max(m, h \cdot a_h)\)\;

  \(\tilde b \leftarrow \mathrm{ToBox}(c_x',c_y',w',h')\)\;

  \If{\(\mathrm{IoU}(\tilde b,b) \ge \tau\)}{
    \Return \(\tilde b\)\;
  }
}
\Return \(b\)\;
\end{algorithm}

\subsubsection{Gaussian Risk-Aware Box Perturbation (GRBP)}

Recent generative vision--language grounding paradigms \cite{qwen2.5vl,kosmos} commonly formulate grounding as a sequence generation problem, where bounding box coordinates are directly predicted as discrete tokens. While this formulation enables end-to-end training, it also introduces significant challenges due to annotation noise and coordinate discretization errors. Under hard supervision, even small geometric deviations between predicted and ground-truth boxes can incur disproportionately large training penalties, leading to unstable optimization and reduced robustness.

To improve the robustness of generative box prediction, we introduce \textbf{Gaussian Risk-Aware Box Perturbation (GRBP)} as a supervision strategy during training. Instead of supervising the model with a single deterministic bounding box target, GRBP applies probabilistic perturbations to the ground-truth box coordinates. Intuitively, larger perturbations correspond to a higher risk of deviating from valid object regions, and vice versa. Based on this observation, we adopt a Gaussian prior to model distributions over box center locations and scales, subject to an intersection-over-union (IoU) constraint with the original box. This formulation produces a set of soft supervision targets that assign higher probability to boxes closer to the ground truth while tolerating small geometric deviations, thereby preserving empirical risk minimization while improving robustness.

Formally, for each ground-truth bounding box \(b\), we generate a perturbed box \(\tilde{b}\) through the following steps:
(i) \textbf{Center Perturbation}: the box center is shifted by Gaussian noise proportional to the box width and height, controlled by a hyperparameter \(\beta\);
(ii) \textbf{Scale Perturbation}: the box width and height are perturbed using multiplicative Gaussian noise, controlled by \(\gamma\), with minimum box size and scale bounds applied to avoid degeneracy;
(iii) \textbf{IoU Guard}: if the perturbed box satisfies \(\mathrm{IoU}(\tilde{b}, b) \ge \tau\), it is accepted; otherwise, the perturbation process is resampled up to \(T\) times, and the original box \(b\) is used as a fallback.
All perturbed boxes are clipped to the image boundaries. The overall procedure is summarized in Algorithm~\ref{algorithm1}.

\subsection{Training Objective and Inference}

\paragraph{Training Objective.}
We train E2E-GMNER using a standard autoregressive maximum likelihood objective over the generated output sequence. Given an image--text pair $(I,T)$ and its corresponding supervision consisting of a reasoning sequence $R$ and structured entity records $\mathcal{Y}$, the training objective is defined as:
\begin{equation}
\mathcal{L} = - \sum_{t} \log p_{\theta}(y_t \mid y_{<t}, \text{Instruction}, I, T),
\end{equation}
where $\{y_t\}$ denotes the tokens in the concatenated output sequence $[R; \mathcal{Y}]$, and $p_{\theta}$ is parameterized by the multimodal large language model.

For entity grounding, bounding box coordinates are serialized as discrete tokens following the output schema in Eq.~\ref{eq:output_schema}. During training, Gaussian Risk-Aware Box Perturbation (GRBP) is applied to generate perturbed bounding boxes, which serve as soft supervision targets for coordinate generation. This noise-aware supervision reduces the sensitivity of the loss to minor geometric deviations while remaining compatible with standard token-level likelihood optimization.

\paragraph{Inference.}
At inference time, E2E-GMNER operates in a fully end-to-end manner without relying on any external teacher models or API-based reasoning supervision. Given a test image--text pair $(I,T)$ and the task instruction, the model autoregressively generates a reasoning sequence followed by structured entity predictions, including entity mentions, semantic types, and grounding bounding boxes.

Importantly, although chain-of-thought reasoning is used as an auxiliary supervision signal during training, no teacher-generated reasoning is required at inference. The model autonomously determines when visual evidence or implicit knowledge is informative and produces final GMNER outputs in a single forward generation pass. This design ensures that inference remains efficient and self-contained, while retaining the benefits of adaptive reasoning learned during instruction tuning.

\begin{table*}[t]
\centering
\footnotesize
\setlength{\tabcolsep}{4pt}
\begin{tabular}{@{}ccccccc@{}}
\toprule
\multirow{2}{*}{Methods} & \multicolumn{3}{c}{Twitter-GMNER} & \multicolumn{3}{c}{Twitter-FMNERG} \\
 & GMNER & MNER & EEG & GMNER & MNER & EEG \\
\midrule
GMDA$^\dagger$ \cite{GMDA} & 58.61 & - & - & 47.37 & - & - \\
GEM$^\dagger$ \cite{GEM} & 59.83 & 83.15 & 63.19 & 50.54 & 68.09 & 63.59 \\
RiVEG$^\dagger$ \cite{RiVEG} & 63.80 & 82.89 & 66.92 & - & - & - \\
MAKAR$^\dagger$ \cite{makar} & 71.88 & 86.38 & 74.64 & 60.54 & 71.24 & 75.66 \\
\midrule
GPT4o \cite{gpt4o} & 41.29 & 65.07 & 44.95 & 32.37 & 52.26 & 41.60 \\
GVATT-OD-EVG \cite{GVATT-OD-EVG} & 48.57 & 76.26 & 53.32 & 40.32 & 60.35 & 54.35 \\
UMT-OD-EVG \cite{UMT-OD-EVG} & 50.29 & 78.58 & 54.78 & 41.32 & 61.63 & 54.43 \\
UMGF-OD-EVG \cite{UMGF-OD-EVG} & 51.67 & 78.83 & 55.74 & 41.92 & 61.79 & 54.75 \\
ITA-OD-EVG \cite{ITA-OD-EVG} & 51.56 & 79.37 & 55.69 & 42.78 & 63.21 & 57.26 \\
MMT5 / BARTMNER-OD-EVG \cite{GMNER} & 52.45 & 80.39 & 55.66 & 45.21 & 66.61 & 58.18 \\
H-Index \cite{GMNER} & 56.41 & 79.73 & 61.18 & 46.55 & 64.84 & 60.46 \\
TIGER \cite{FGMNER} & 57.48 & - & - & 47.20 & 64.91 & 61.96 \\
MQSPN \cite{MQSPN2025AAAI} & 58.76 & 80.43 & 62.40 & 47.86 & 66.83 & 61.95 \\
UnCo \cite{unco} & 58.83 & 79.55 & 63.49 & 48.17 & 65.06 & 62.73 \\
\midrule
E2E-GMNER-7B(Ours) & 63.94 & 77.65 & 66.12 &  54.32 & 65.67 & 66.78 \\
\bottomrule
\end{tabular}
\caption{Performance comparison of different methods on Twitter-GMNER and Twitter-FMNERG datasets. $^\dagger$ indicates the methods using additional data or knowledge augmentation.}
\label{tab:results}
\end{table*}

\section{Experiment}
\subsection{Datasets}
We evaluate on two social-media Grounded Multimodal NER benchmarks: \textbf{Twitter-GMNER} \cite{GMNER} and \textbf{Twitter-FMNERG} \cite{FGMNER}. Both datasets annotate image–text pairs with entity spans, types, and grounding regions in the image. Twitter-GMNER provides four coarse-grained entity types, while Twitter-FMNERG expands the label space to eight coarse-grained types and 51 fine-grained subtypes for more detailed evaluation. 

\subsection{Evaluation Metrics}
Following \cite{GMNER}, we report F1 for the overall \textbf{GMNER} task and its two subtasks: \textbf{Multimodal Named Entity Recognition (MNER)} and \textbf{Entity Extraction and Grounding (EEG)}. MNER evaluates span and type correctness, while EEG evaluates span extraction with grounding correctness under the standard intersection-over-union criterion. The overall GMNER score requires both correct recognition and correct grounding. All formal definitions and formulas are deferred to Appendix~\ref{sec:appendix-formulas}.

\subsection{Baselines}
To evaluate our framework, we compare against three categories of baselines. The first category includes methods that leverage external knowledge or additional data beyond the standard training set. Specifically, GMDA$^\dagger$ \cite{GMDA}, GEM$^\dagger$ \cite{GEM}, RiVEG$^\dagger$ \cite{RiVEG}, and MAKAR$^\dagger$ \cite{makar} incorporate auxiliary resources—such as pre-trained vision-language models, extra datasets, or structured knowledge—to enhance multimodal entity recognition and grounding.

The second category consists of pipeline-based approaches, and we further divide it into two subgroups. The first subgroup (GPT4o \cite{gpt4o}, GVATT-OD-EVG \cite{GVATT-OD-EVG}, UMT-OD-EVG \cite{UMT-OD-EVG}, UMGF-OD-EVG \cite{UMGF-OD-EVG}, ITA-OD-EVG \cite{ITA-OD-EVG}, and MMT5/BARTMNER-OD-EVG \cite{GMNER}) follows an MNER-first paradigm: they first extract textual entities using a named entity recognition model and then ground these entities to visual regions via object detectors or alignment modules. The second subgroup (H-Index \cite{GMNER}, TIGER \cite{FGMNER}, MQSPN \cite{MQSPN2025AAAI}, and UnCo \cite{unco}) adopts a feature-fusion strategy: they independently encode text with a language encoder and extract salient visual regions using an object detector, then fuse these representations through cross-modal interaction layers to jointly predict entities and their grounded spans. Despite their strong performance, both subgroups suffer from limited end-to-end optimization and error propagation between stages.

Finally, we include our proposed models, which are fully end-to-end generative frameworks. Unlike prior work, they jointly generate textual entities and their corresponding visual bounding boxes in a single sequence without relying on external resources or intermediate pipelines, enabling tighter cross-modal alignment and more robust grounding.

\subsection{Main result}
In Table \ref{tab:results}, we compare the performance of our method with several state-of-the-art approaches under three evaluation metrics: GMNER, MNER, and EEG, on the Twitter-GMNER and Twitter-FMNERG datasets. The compared baselines can be broadly categorized into two groups: methods that rely on external knowledge sources and those that do not. Overall, the experimental results show that our proposed E2E-GMNER consistently outperforms all state-of-the-art methods that do not leverage external knowledge, and achieves competitive performance even when compared with methods that explicitly incorporate external knowledge sources. Based on these results, we draw the following conclusions.

(1) As the first fully end-to-end framework for GMNER, E2E-GMNER outperforms all pipeline-based methods except MAKAR, which relies on external knowledge sources.\footnote{This method may be less cost-effective in practice, as it requires full-parameter fine-tuning on 8×A100 GPUs. In addition, its reliance on an external search engine necessitates internet access.} This demonstrates that jointly modeling entity recognition, visual grounding, semantic understanding, and knowledge reasoning in a unified generative framework effectively mitigates error accumulation in cascaded architectures and enables more effective joint optimization.

(2) Compared with knowledge-augmented approaches, E2E-GMNER still achieves highly competitive performance, indicating that our method effectively exploits the implicit knowledge acquired by multimodal large language models during pretraining and successfully adapts it to the GMNER task without requiring explicit external knowledge retrieval.

\begin{table}[t]
\centering
\begin{tabular}{lcc}
\toprule
\textbf{Method} & \textbf{GMNER} & \textbf{FMNERG} \\
\midrule
E2E-GMNER         & 63.94          & 54.32          \\
\midrule
w/o CoT           & 62.57          & 53.96          \\
w/o GRBP          & 61.51          & 52.92          \\
\bottomrule
\end{tabular}
\caption{Ablation study of E2E-GMNER on GMNER and FMNERG datasets.}
\label{tab:ablation}
\end{table}

\subsection{Ablation Study}

To assess the contribution of the key components in our end-to-end generative framework, we conduct an ablation study by individually removing Chain-of-Thought (CoT) reasoning and Gaussian Risk-Aware Box Perturbation (GRBP). The results are summarized in Table~\ref{tab:ablation}.

Removing CoT reasoning (\textit{w/o CoT}) leads to a clear and consistent performance degradation on both GMNER and FMNERG. This indicates that CoT reasoning plays a critical role in enabling the model to perform structured multimodal reasoning, particularly in determining when visual evidence or implicit background knowledge should be leveraged. Without this reasoning mechanism, the model is more prone to relying on noisy or irrelevant visual cues, which adversely affects its ability to jointly perform entity recognition and grounding.

Eliminating GRBP (\textit{w/o GRBP}) also results in a noticeable performance drop, though to a lesser extent compared with removing CoT. This observation suggests that GRBP effectively improves the robustness of generative grounding by alleviating the sensitivity of discrete bounding box prediction to annotation noise and discretization errors. By providing noise-aware supervision, GRBP helps stabilize training and improves generalization in multimodal grounding scenarios.

\begin{table}[t]
\centering
\begin{tabular}{lcc}
\toprule
\textbf{Teacher Prompt} & \textbf{GMNER} & \textbf{FMNERG} \\
\midrule
Qwen2.5-VL-72B       & 63.94          & 54.32         \\
Qwen2.5-VL-7B       & 61.23          & 52.74         \\
GPT4o      & 63.77          & 53.96           \\
\bottomrule
\end{tabular}
\caption{Performance comparison of different teacher prompts in generating CoT data on GMNER and FMNERG datasets.}
\label{tab:teacher_prompt}
\end{table}

\begin{figure*}[t]
  \centering
  \includegraphics[width=\linewidth]{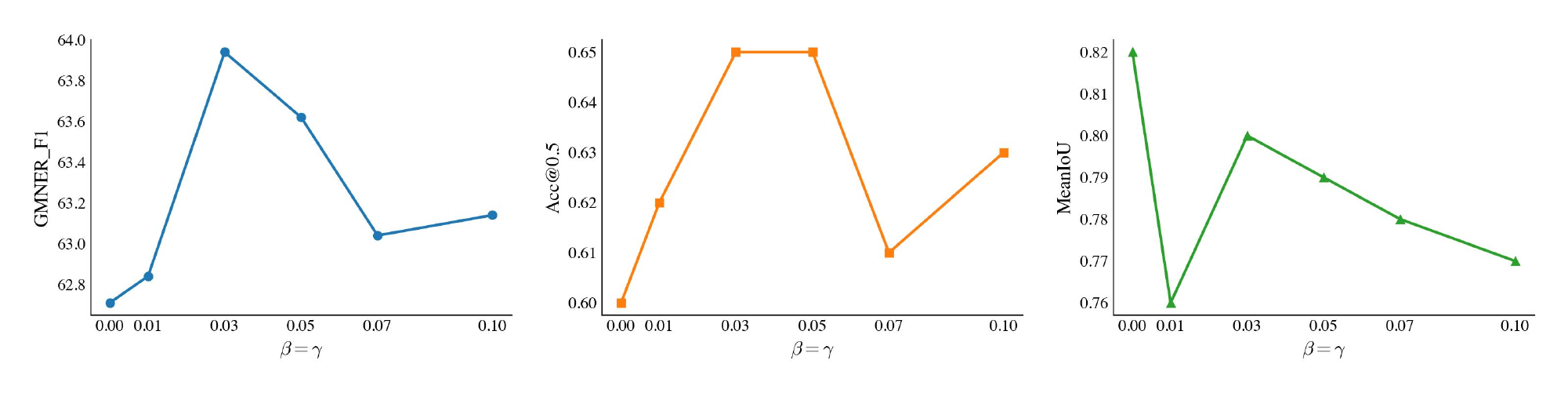}
  \caption{Impact of Gaussian perturbation strength on GMNER performance. The figure reports Acc@0.5, GMNER F1 score, and MeanIoU under different values of the Gaussian perturbation factors $\beta$ and $\gamma$, illustrating the effect of Gaussian Risk-Aware Box Perturbation (GRBP) on entity grounding recall, overall extraction performance, and localization accuracy.}
  \label{figure3}
\end{figure*}

\subsection{Impact of Teacher Prompts on CoT Data Generation}

To investigate how the choice of teacher model affects the quality of chain-of-thought (CoT) supervision and the resulting GMNER performance, we compare different teacher prompts used during CoT data generation, as summarized in Table~\ref{tab:teacher_prompt}. The results indicate that larger teacher models generally lead to better downstream performance, suggesting that increased model capacity enables the generation of more informative and reliable reasoning traces. In addition, although GPT4o produces competitive results, it underperforms compared with the Qwen2.5-VL-72B series. A plausible explanation is that our finetuned backbone model belongs to the Qwen-2.5-VL family, whose reasoning patterns and multimodal representations are more consistent with those of the Qwen-based teachers, resulting in better alignment during instruction tuning. Finally, it is worth noting that all evaluated teacher models yield reasonably strong performance, demonstrating that our training strategy is robust and effective across different teacher choices, rather than being tightly coupled to a specific teacher model.

\subsection{Perturbation Strength Analysis}

We conduct a systematic analysis of different Gaussian perturbation strengths to investigate the impact of Gaussian Risk-Aware Box Perturbation (GRBP) on both entity grounding and overall GMNER performance, as shown in Figure~\ref{figure3}. Specifically, we compare models trained with varying values of the perturbation factors $\beta$ and $\gamma$ in terms of three metrics: Acc@0.5, which reflects the recall of correctly grounded entities with valid bounding boxes; the overall GMNER F1 score; and the MeanIoU between predicted bounding boxes and gold annotations.

The results show that a moderate perturbation strength achieves the best overall performance, with $\beta = \gamma = 0.03$ yielding the most favorable trade-off across metrics. Compared with training without perturbation, the GMNER F1 score exhibits only a marginal difference, while Acc@0.5 shows a substantially larger improvement. This indicates that GRBP primarily enhances the recall of the grounding process, enabling the model to correctly associate more entities with valid visual regions, even when the predicted boxes are not perfectly aligned with the gold annotations.

Interestingly, under $\beta = \gamma = 0.03$, the MeanIoU between predicted and gold bounding boxes slightly decreases compared with the no-perturbation setting. However, this minor reduction in localization precision is accompanied by a clear improvement in grounding recall and overall extraction performance. This observation suggests that GRBP encourages the model to be more tolerant of small geometric deviations during training, which improves its ability to generalize and recover valid grounding regions, rather than overfitting to exact box coordinates.

Overall, this analysis shows that GRBP provides a favorable robustness–precision trade-off: moderate Gaussian perturbations slightly reduce localization accuracy in terms of MeanIoU, but substantially improve grounding recall, resulting in better end-to-end GMNER performance.

\section{Conclusion}
In this work, we presented E2E-GMNER, the first fully end-to-end generative framework for grounded multimodal named entity recognition that jointly models entity extraction, semantic classification, and visual grounding within a single vision–language model. By eliminating cascaded pipeline components, our approach enables unified optimization and mitigates error accumulation inherent in prior methods. The incorporation of chain-of-thought instruction tuning allows the model to adaptively leverage visual cues and implicit knowledge only when beneficial, while Gaussian Risk-Aware Box Perturbation improves the robustness and stability of generative grounding under annotation noise. Experimental results on two widely used GMNER benchmarks show that E2E-GMNER consistently outperforms or matches strong baselines without relying on external object detectors or online knowledge sources. These findings highlight the promise of end-to-end generative paradigms for structured multimodal understanding and suggest future directions for extending such frameworks to broader multimodal information extraction tasks.

\section*{Limitations}

Although E2E-GMNER demonstrates strong performance on grounded multimodal named entity recognition, our current framework is specifically designed and evaluated for the GMNER task. The model architecture, output schema, and training objectives are tailored to jointly predict entity spans, semantic types, and grounding boxes, and have not yet been adapted or validated on other multimodal information extraction tasks such as relation extraction or event extraction. In addition, our approach relies on chain-of-thought supervision during training, which introduces extra annotation or teacher-model inference cost. Exploring task-agnostic formulations and reducing the dependence on auxiliary reasoning supervision remain important directions for future work.

\section*{Acknowledgement}
This work was supported by the Young Scientists Fund of the National Natural Science Foundation of China (NSFC) [Grant No. 62506058]

\bibliography{custom}

\appendix

\section{Formulas and Metrics}
\label{sec:appendix-formulas}
\begin{align}
C_{e,t} &= 
\begin{cases} 
1 & \text{if } \hat{e}=e \land \hat{t}=t, \\
0 & \text{otherwise}
\end{cases} \\
C_{b} &= 
\begin{cases} 
1 & \text{if } \hat{b}=\varnothing \land b=\varnothing \lor \mathrm{IoU}(\hat{b},b)\ge 0.5, \\
0 & \text{otherwise}
\end{cases}
\end{align}

For \textbf{MNER}, a prediction is correct if and only if \( C_{e,t}=1 \).

For \textbf{EEG}, a prediction is correct if and only if \( \hat{e}=e \) and \( C_{b}=1 \).

For \textbf{MNERG}, a prediction is correct if and only if

\begin{equation}
C = C_{e,t} \cdot C_{b} = 1.
\label{eq:triplet_correct}
\end{equation}

Let \(\#\mathrm{correct}\), \(\#\mathrm{pred}\), and \(\#\mathrm{gold}\) represent the number of correct predictions, total predictions, and gold records, respectively. Precision, recall, and F1 score are computed as:

\begin{equation}
\mathrm{Pre}=\frac{\#\mathrm{correct}}{\#\mathrm{pred}},\qquad
\mathrm{Rec}=\frac{\#\mathrm{correct}}{\#\mathrm{gold}},
\label{eq:pr}
\end{equation}

\begin{equation}
\mathrm{F1}=\frac{2\cdot \mathrm{Pre}\cdot \mathrm{Rec}}{\mathrm{Pre}+\mathrm{Rec}}.
\label{eq:f1}
\end{equation}

In addition to the standard metrics, we also report the following:

The accuracy at specific IoU thresholds (e.g., 0.5 and 0.75) is used to evaluate the model's performance in extracting and grounding entities. Let \( N_{\text{matched}} \) denote the number of matched pairs at a given threshold. The accuracy at IoU threshold \( \mathrm{IoU_{thr}} \) is computed as:

\begin{equation}
\text{Acc@IoU}_{\text{thr}} = \frac{N_{\text{matched}}}{\#\mathrm{gold\ with\ box}}.
\label{eq:acc_threshold}
\end{equation}

The overall mean IoU is the average IoU between all gold entity boxes and their respective predicted boxes. It is computed as:

\begin{equation}
\text{Overall Mean IoU} = \frac{\sum_{i=1}^{N_{\text{gold}}} \mathrm{IoU}(b_i, \hat{b}_i)}{N_{\text{gold}}}.
\label{eq:overall_miou}
\end{equation}

\section{Implementation Details}
All experiments are conducted on a single NVIDIA RTX 5090 GPU. To construct Chain-of-Thought (CoT) training data, we leverage three strong vision-language teacher models: \texttt{Qwen2.5-VL-72B-Instruct}, \texttt{Qwen2.5-VL-7B-Instruct}, and \texttt{GPT-4o}. Our primary experiments fine-tune the  \texttt{Qwen2.5-VL-7B} vision-language models as backbones. We optimize using AdamW with a learning rate of $4 \times 10^{-5}$ and weight decay of $1 \times 10^{-3}$. The per-device batch size is set to 2, and we apply gradient accumulation over 8 steps, resulting in an effective batch size of 16. The maximum input sequence length is limited to 2048 tokens. To improve training efficiency and reduce memory footprint, we employ Low-Rank Adaptation (LoRA) during fine-tuning.

\end{document}